# LSENet: Location and Seasonality Enhanced Network for Multi-Class Ocean Front Detection

Cui Xie, *Member, IEEE*, Hao Guo, and Junyu Dong, *Member, IEEE*

*Abstract*—Ocean fronts can cause the accumulation of nutrients and affect the propagation of underwater sound, so high-precision ocean front detection is of great significance to the marine fishery and national defense fields. However, the current ocean front detection methods either have low detection accuracy or most can only detect the occurrence of ocean front by binary classification, rarely considering the differences of the characteristics of multiple ocean fronts in different sea areas. In order to solve the above problems, we propose a semantic segmentation network called location and seasonality enhanced network (LSENet) for multi-class ocean fronts detection at pixel level. In this network, we first design a channel supervision unit structure, which integrates the seasonal characteristics of the ocean front itself and the contextual information to improve the detection accuracy. We also introduce a location attention mechanism to adaptively assign attention weights to the fronts according to their frequently occurred sea area, which can further improve the accuracy of multi-class ocean front detection. Compared with other semantic segmentation methods and current representative ocean front detection method, the experimental results demonstrate convincingly that our method is more effective.

## I. Introduction

OCEAN fronts refer to the narrow transition zone between two or more water masses of different physical properties, which can be described by the horizontal gradient of sea temperature, salinity, density, chlorophyll and other marine hydrological elements. In ocean fronts zone, the strong local horizontal and vertical flows lead to the accumulation of nutrients and plankton. Therefore, the fronts have great influences on mariculture and pelagic fishing planning. In addition, ocean fronts have obvious reflection and refraction effect on the propagation of underwater sound, which will affect the marine military and other underwater operations. Thus, it is of great significance and value to identify the ocean front.

Traditional ocean front detection methods are mainly based on gradient algorithm [1-3]. These methods heavily depend on the experience of ocean experts to determine the gradient threshold, which is unstable, time-consuming and not automatic. Moreover, these methods are seriously affected by noise data, so the detection accuracy is low. In recent years, with the development of deep learning, some ocean front detection methods based on neural network have been proposed. Sun *et al.* [4] proposed an ocean front detection method based on convolutional neural network, but the algorithm could only detect whether the ocean front occurs (binary classification) at patch level accuracy. Li *et al.* [5] regarded ocean front detection as an edge detection problem and proposed the weak edge identification network (WEIN), which can identify the existence of the ocean front at pixel level. However, all the above ocean front detection methods ignore the detection of different classes of ocean fronts in different sea areas. In practical research, researchers often focus on a specific class of ocean front in specific sea area, or compare the differences between different classes of ocean fronts. Therefore, the detection of different classes of ocean fronts is urgently needed for more comprehensive analysis. Although Cao *et al.* [6] have studied the detection of multi-class ocean fronts. Their proposed method is still based on gradient algorithm, the detection accuracy and generalization ability of the method are not satisfactory.

In order to solve the above problems, we propose a new semantic segmentation network model called location and seasonality enhanced network (LSENet). In addition, a channel supervision unit structure and a location attention mechanism are designed to integrate the seasonality and spatial dependence of ocean fronts, so as to improve the detection accuracy of the model. The main contributions of this paper are as follows:

1) We model multi-class detection of ocean fronts as a semantic segmentation problem and propose LSENet to address it.

2) We propose a channel supervised unit structure, which integrates the seasonality of ocean front to the feature learning process of the model.

3) We propose a new location attention mechanism, which enables the model to adaptively allocate the weights to ocean fronts in different sea areas, thus improving the multi-class detection accuracy of the model.

The remaining part of this paper is organized as follows. Section II provides an overview of the related work. Section III describes the data sets we used. Section IV describes the details of the LSENet model. Experiments and some visualization

This work was supported in part by the National Key Research and Development Program of China under Grant 2020YFE0201200, the National Natural Science Foundation of China (NSFC) under Grant 41706010, U1706218, and 41927805, and the Fundamental Research Funds for the Central Universities under Grant 201964022 (Corresponding authors: Junyu Dong).

C. Xie, H. Guo and J. Dong are with the School of Computer Science and Technology, Ocean University of China, Qingdao 266100, China (E-mail: spring@ouc.edu.cn (C. Xie), dongjunyu@ouc.edu.cn (J. Dong)).

results are presented in section V. Section VI concludes the paper.

## II. Related Work

### A. Semantic Segmentation

Currently, semantic segmentation is one of the hot research topics in the field of computer vision. Its goal is to classify every pixel in the image [7]. It has been widely used in biomedical image segmentation [8], automatic driving [9] and other fields. Long *et al.* [10] proposed FCN (Fully Convolutional Networks) in 2015, which is considered as a pioneering work of semantic segmentation. Only convolutional neural network is used in this network model, and information fusion of different scales is added. Ronneberger [11] *et al.* proposed UNet, which performs well in the case of weak supervised learning, and is still the common baseline in the field of biomedical image segmentation. Zhao *et al.* [12] proposed PSPNet on the basis of FCN, and achieved good performance in some semantic segmentation data sets (such as Cityscapes dataset [13]) by introducing pyramid pooling module. In addition, the Deeplab series models proposed by Chen *et al.* [14-17] performed well in semantic segmentation tasks, their Xception structure significantly reducing the number of parameters and improving the speed of model operation. This method is also one of the most frequently chosen baselines in the field of semantic segmentation. Based on the successful application of semantic segmentation in many fields, we introduce the idea of semantic segmentation into ocean front detection problem to accomplish the task of multi-class ocean fronts detection at pixel level.

### B. Attention Mechanisms

In recent years, attention mechanism has been widely used in the field of semantic segmentation. Hu *et al.* [18] proposed a kind of channel attention in their SENet, which can help models to capture global semantic information at the channel level. Fu *et al.* [19] proposed channel correlation attention and position correlation attention in order to capture the correlation between pixels. Choi *et al.* [20] found that in urban-scene image, the sky basically appeared in the upper part of the image, while the road usually existed in the lower part of the image. Therefore, they proposed a height attention, which enabled the model to capture the relationship between height and classes and improved the segmentation accuracy. Inspired by the work of Choi *et al.* [20], we propose a location attention that aims to ensure that the model learns the spatial dependence between the ocean front classes and the corresponding sea areas. In other words, it makes sure the model knows that a front occurring near the Bohai Sea in China is more likely to be a Bohai Coastal Front than a Korean Strait Front.

## III. Datasets

It is worth mentioning that there is no multi-class labeled data in the current ocean fronts research. So, we built a dataset of ocean fronts to train the model. Firstly, we collected the advanced high resolution radiometer (AVHRR) sea surface

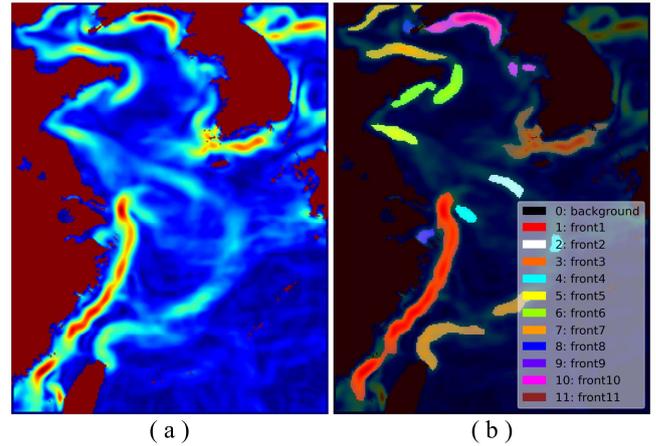
Fig. 1. Labeled data. (a) original SST gradient map. (b) labeled result.

temperature (SST) daily data (2015 to 2018) from National Oceanic and Atmospheric Administration (NOAA). Then, we intercepted a piece of data near the sea area of China, with longitude ranging from 117.975E to 130.975E and latitude ranging from 23.025N to 40.025N. Next, we computed the gradient of daily SST data and generate 1461 SST gradient maps of $340 \times 260$ size. Finally, we invited experts in ocean fronts research to help label these SST gradient maps, and the labeled results are shown in Fig. 1. Here, according to the experts' experience and the existing research [21]-[24], ocean fronts near sea area of China were divided into 11 classes, namely: (1) Zhejiang-Fujian Front (2) Yellow Sea Warm Current Front (3) Kuroshio Front (4) Northern East China Sea Shelf Front (5) Jiangsu Front (6) Shandong Peninsula Front (7) Bohai Sea Coastal Front (8) Bohai Sea Strait Front (9) Yangtze Estuary Front (10) Western Korea Coastal Front (11) Korea Strait Front.

In order to evaluate the performance of the model comprehensively and accurately, we selected the data of the whole year of 2018 (365 samples in total) as the test set to ensure that the test set contained ocean fronts from all seasons of the year. We used the remaining data from 2015 to 2017 (a total of 1096 samples) as the train set for model training. In order to prevent overfitting and increase the amount of data at the same time, we used random photometric distortion, random crop and random flip image to enhance the data. Moreover, in order to facilitate data enhancement and model downsampling, we used 0 padding to reshape the size of the original input image to $352 \times 352 \times 3$.

## IV. Model Description

### A. Overview

Fig. 2 shows the overview of our model. It consists of two parts: (1) encoder-decoder part, which is responsible for feature extraction and feature processing; (2) head part, which is responsible for output detection results. The input image will first go through 5 encoders and 4 decoders. The role of the encoder is to learn the features of the image, and downsampling the image, so as to increase the receptive field of the convolution kernel and reduce the computation of the network.

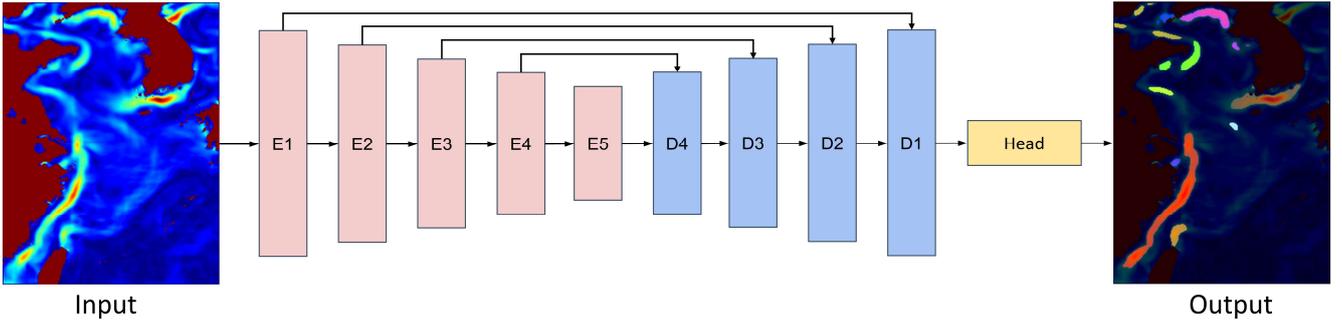

Input  Output
Fig. 2. An overview of the LSENet.

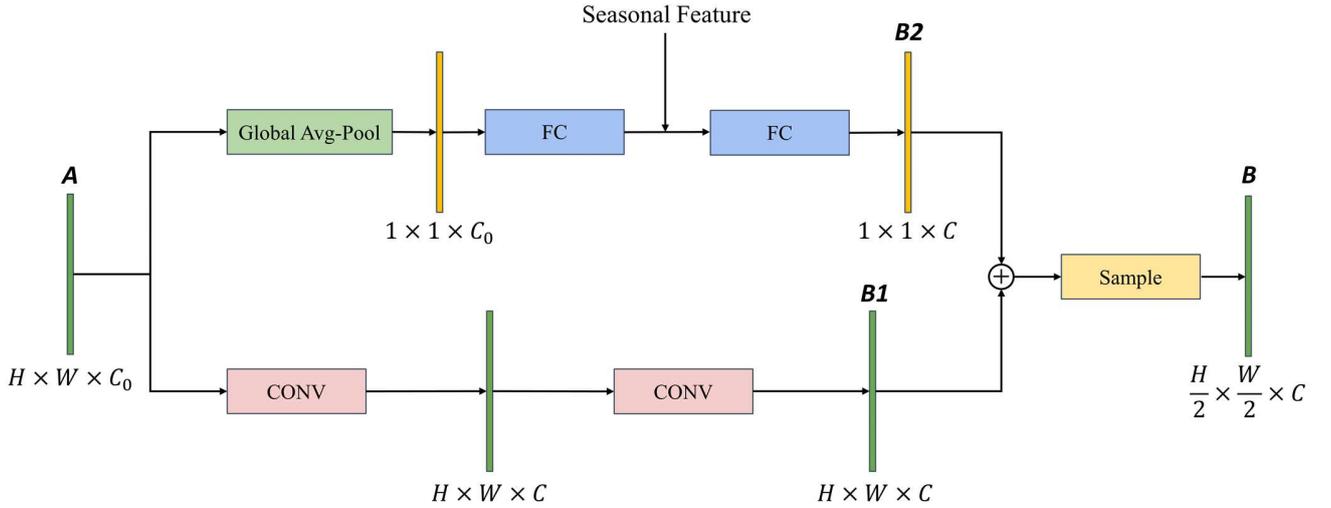

Fig. 3. The structure of an encoder.

The function of the decoder is to analyze the feature map generated by the encoder, and to upsample the feature map step by step to the original input size. Moreover, in order to reduce the loss of image information caused by the downsampling, the concatenate operation is added between the encoder and the decoder. Finally, the multi-class detection results are generated through the head part.

*B. Encoder-Decoder Part*

Fig. 3 shows the structure of an encoder. The encoder contains two branches. The branch at the bottom is the image feature processing branch, which is composed of two convolution blocks. Its main function is to extract and analyze the features of the image. In this branch, the feature map $\mathbf{A} \in \mathbb{R}^{H \times W \times C_0}$ are fed into two convolution blocks to generate a new feature map $\mathbf{B1} \in \mathbb{R}^{H \times W \times C}$. The upper branch is the channel supervision unit proposed in this paper. The design idea of channel supervision unit is to let the model comprehensively analyze the input feature map at the channel level, judge the class of fronts contained in the feature map, and introduce the seasonality of ocean fronts into the model in the form of seasonal feature encoding. In this branch, feature map $\mathbf{A} \in \mathbb{R}^{H \times W \times C_0}$ first goes through a global average pooling layer and the dimension is compressed to $\mathbb{R}^{1 \times 1 \times C_0}$. Then a full connection layer (RELU function activation) is used to reduce the number of channels to $C/2$ for seasonal feature insertion. After concatenating the monthly one-hot encoded seasonal features, the number of channels is changed to $12 + C/2$, and then a full connection layer (RELU function activation) is used to generate the feature map $\mathbf{B2} \in \mathbb{R}^{1 \times 1 \times C}$. After summing the feature map $\mathbf{B2}$ with each pixel of $\mathbf{B1}$, the output feature map $\mathbf{B} \in \mathbb{R}^{\frac{H}{2} \times \frac{W}{2} \times C}$ of the encoder is obtained by downsampling. The structure of the decoder and encoder is basically the same, the only difference is that the last sampling layer of decoder uses upsampling. All encoder and decoder parameters used in our model are shown in Table I.

*C. Head Part*

The head part consists of detection branch and attention branch (as shown in Fig. 4). The detection branch at the bottom is relatively simple, which only contains a convolutional layer of 1×1 convolution kernel to map the number of channels to the desired number of classes $N$ ($N = 12$), and then the model detection result is obtained by Softmax function. For the upper location attention branch, a given feature map $\mathbf{A} \in \mathbb{R}^{H \times W \times C}$ is first fed into an averaged pooling layer to obtain the feature map $\mathbf{C1} \in \mathbb{R}^{\frac{H}{r} \times \frac{W}{r} \times C}$, where $r$ is the downsampling coefficient, which is empirically set to 11 in this paper. Under this value, the size of feature map $\mathbf{C1}$ is 32 × 32. Then, the feature map $\mathbf{C1}$ is fed into a convolution block for feature learning, and the size of the feature map is reshaped to $\mathbb{R}^{\frac{H}{r} \times \frac{W}{r} \times C_l}$. Each pixel in this feature map can be considered as a high abstraction of the feature of the region with the size of $r \times r$. At the same time, in order to enhance the model's sensitivity to geographical location, location encoding is added into the attention branch. Next, the location encoded feature map is input into several convolution blocks for feature learning, and then the result is converted into values between 0 and 1 through a convolution block with Sigmoid activation function. These values represent the attention weights of the model for different channels of each region. The resulting attention weight map $\mathbf{C2} \in \mathbb{R}^{\frac{H}{r} \times \frac{W}{r} \times N}$ is upsampled by bilinear interpolation and is element-wise multiplied with the results of the detection branch. Finally, an element-wise sum operation is performed with the result of multiplication and the result of detection branch to further optimize the detection result.

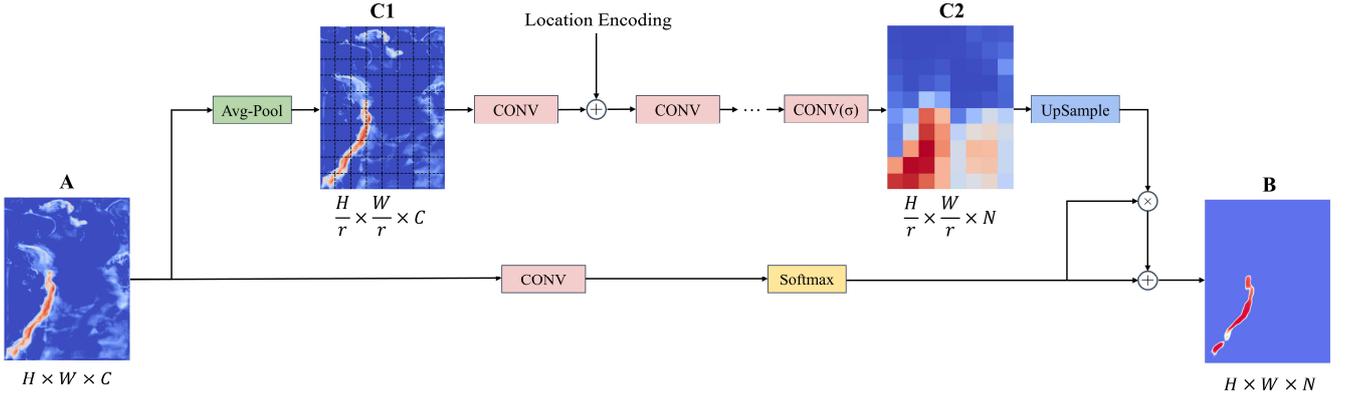

Fig. 4. The structure of the head part.

TABLE I
PARAMETERS IN ENCODERS AND DECODERS

| Block | CONV | FC | Sample |
|---|---|---|---|
| E1 | conv, 3×3, 64<br>conv, 3×3, 64 | fc, 32<br>fc, 64 | max pool, 2×2 |
| E2 | conv, 3×3, 128<br>conv, 3×3, 128 | fc, 64<br>fc, 128 | max pool, 2×2 |
| E3 | conv, 3×3, 256<br>conv, 3×3, 256 | fc, 128<br>fc, 256 | max pool, 2×2 |
| E4 | conv, 3×3, 512<br>conv, 3×3, 512 | fc, 256<br>fc, 512 | max pool, 2×2 |
| E5 | conv, 3×3, 1024<br>conv, 3×3, 1024 | fc, 512<br>fc, 1024 | max pool, 2×2 |
| D4 | conv, 3×3, 512<br>conv, 3×3, 512 | fc, 256<br>fc, 512 | nearest neighbor interpolate |
| D3 | conv, 3×3, 256<br>conv, 3×3, 256 | fc, 128<br>fc, 256 | nearest neighbor interpolate |
| D2 | conv, 3×3, 128<br>conv, 3×3, 128 | fc, 64<br>fc, 128 | nearest neighbor interpolate |
| D1 | conv, 3×3, 64<br>conv, 3×3, 64 | fc, 32<br>fc, 64 | nearest neighbor interpolate |

The location encoding uses positional encoding 2D method of wang et al. [25], which can not only encode location order from left to right, but also top to bottom. And the formula is as follows:

$$PE(x, y, 2i) = sin(x/10000^{4i/D}) \quad (1)$$
$$PE(x, y, 2i + 1) = cos(x/10000^{4i/D}) \quad (2)$$
$$PE(x, y, 2j + D/2) = sin(y/10000^{4j/D}) \quad (3)$$
$$PE(x, y, 2i + 1 + D/2) = cos(y/10000^{4j/D}) \quad (4)$$

Where, $x$ and $y$ represent the values of horizontal and vertical coordinates respectively, and the value ranges are $[0, H]$, $[0, W]$. $i$ and $j$ represent the number of channels whose value range is $[0, D/2]$. $D$ represents the total number of channels in the feature map. The location encoding is completed by element-wise sum the result of the positional encoding 2D and the feature map to be encoded.

V. EXPERIMENTS

*A. Parameter Setting and Evaluation Metrics*

Based on our generated datasets in Part III, all experiments in this paper are performed on an NVIDIA RTX 3080 GPU and implemented using the open source deep learning library Keras [26] and TensorFlow [27]. In order to compare with other semantic segmentation models, we set the same parameters for our model and the compared model: batch size is set to 4, training epoch is set to 80, cross entropy is used as loss function, Adam [28] is used as optimizer, and 1e-3 is used as initial learning rate. And the learning rate decay strategy is used to reduce the learning rate to half of the previous rate when the loss of the validation set does not drop for 3 consecutive epochs.

In this paper, we use mean intersection over union (mIoU) as an evaluation metric. The mIoU is a commonly used evaluation method in the field of semantic segmentation. The essence of intersection over union (IoU) is to calculate the ratio of the intersection and the union of the ground truth and the predicted segmentation datasets. This ratio can be expressed as true positive (TP) divided by the sum of false positive (FP), false negative (FN), and true positive (TP), as the following formula:

$$IoU = TP/(FP + FN + TP) \quad (5)$$

The mIoU is the average of IoU, and the complete formula is as follows:

$$mIoU = \frac{1}{k+1} \sum_{i=0}^{k} \frac{p_{ii}}{\sum_{j=0}^{k} p_{ij} + \sum_{j=0}^{k} p_{ji} - p_{ii}} \quad (6)$$

TABLE II
COMPARISON OF MIOU WITH OTHER SEMANTIC SEGMENTATION MODELS

| Model | background | front1 | front2 | front3 | front4 | front5 | front6 | front7 | front8 | front9 | front10 | front11 | mIoU |
|---|---|---|---|---|---|---|---|---|---|---|---|---|---|
| FCN-8s | 97.70 | 73.22 | 60.03 | 61.14 | 28.87 | 43.26 | 57.25 | 56.47 | 46.25 | 32.24 | 67.37 | 70.41 | 57.85 |
| UNet | 98.56 | **80.32** | 66.74 | 68.65 | 29.38 | 57.11 | 73.51 | 74.73 | 64.62 | 47.33 | 83.46 | 84.34 | 69.06 |
| SegNet | 97.84 | 74.10 | 57.43 | 53.26 | 24.75 | 45.67 | 58.84 | 63.12 | 47.18 | 32.90 | 72.52 | 67.68 | 57.94 |
| DeepLabv3+ | 98.36 | 73.26 | 69.06 | 69.66 | 29.13 | 56.53 | 67.53 | 71.33 | 60.53 | 37.20 | 80.74 | 81.84 | 66.27 |
| PSPNet (ResNet50) | 97.85 | 72.42 | 66.43 | 62.86 | 30.11 | 46.72 | 63.52 | 56.99 | 45.34 | 33.92 | 70.84 | 72.86 | 59.99 |
| PSPNet (ResNet101) | 97.93 | 72.49 | 65.22 | 65.27 | 29.87 | 51.96 | 64.81 | 59.73 | 46.63 | 34.81 | 71.17 | 74.24 | 61.18 |
| DANet (ResNet50) | 98.57 | 79.53 | 67.06 | 69.12 | 31.64 | 57.11 | 73.98 | 75.12 | 66.77 | **49.60** | 83.18 | 83.55 | 69.60 |
| DANet (ResNet101) | 98.57 | 78.96 | 70.48 | 69.58 | 26.73 | 57.86 | **74.56** | 74.59 | **68.68** | 44.81 | 82.75 | 84.05 | 69.30 |
| HANet | 98.37 | 73.84 | 68.63 | 70.77 | 25.56 | 56.93 | 68.77 | 71.32 | 59.28 | 32.61 | 80.69 | 81.39 | 65.68 |
| Ours | **98.59** | 79.22 | **73.54** | **71.15** | **38.43** | **60.59** | 72.89 | **75.53** | 66.90 | 48.78 | **84.66** | **84.70** | **71.25** |

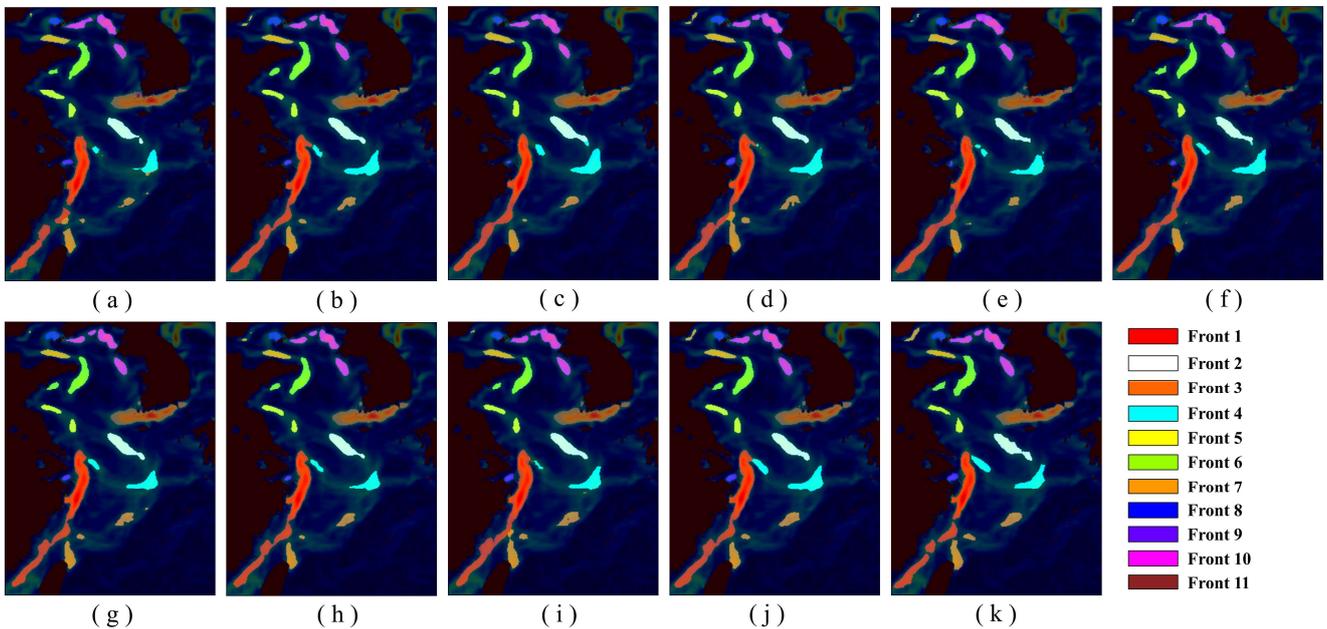

Fig. 5. Comparison with other semantic segmentation models. (a) FCN-8s. (b) UNet. (c) SegNet. (d) Deeplabv3+. (e) PSPNet (ResNet50). (f) PSPNet (ResNet101). (g) DANet (ResNet50). (h) DANet (ResNet101). (i) HANet. (j) our proposed model. (k) ground truth.

Where, $p_{ij}$ represents the number of pixels that belong to class $i$ but are detected to be class $j$ ($p_{ij} = TP$, when $i$ and $j$ are equal, and $p_{ij} = FN$, when $i$ and $j$ are not equal). Similarly, $p_{ji}$ represents the case of FP when $i$ and $j$ are not equal. And $k$ represents the total number of classes.

*B. Comparison with Other Semantic Segmentation Models*

In order to evaluate the results of model detection, some semantic segmentation models are used for comparison:

1) **FCN-8s** [10] is a fully convolutional network without fully connected layer, with the introduction of deconvolutional layer to increase data size, which can output fine results.

2) **UNet** [11] uses encoder-decoder structure, it performs well in weakly supervised learning and is widely used in the field of biomedical image segmentation.

3) **SegNet** [29] uses max-pooling indices that can record the location information of maximum pixels, which solves the problem of information loss.

4) **Deeplabv3+** [16] uses the structure of dilated convolution and Xception to improve speed of the algorithm and performs well in the field of semantic segmentation.

5) **PSPNet** [12] uses pyramid pooling module to aggregate background information and is widely used in the field of semantic segmentation.

6) **DANet** [18] is able to capture the correlation between pixels by using the two attention mechanisms of channel correlation and position correlation.

7) **HANet** [19] adds height attention based on Deeplabv3+, which enables the model to capture the relationship between height and classes.

In addition, PSPNet and DANet use ResNet50 [30] and ResNet101 [30] as backbone respectively to process and learn features.

Table II shows the comparison results, from which we can see that our model does perform well, especially for the detection of front4 with variety of morphology. Although the performance of FCN and PSPNet are good for some applications, they don't perform well for fronts detection. Because both methods have more than 2× upsampling layer, which will cause information loss for detection of strip-shaped fronts. In addition, performance of DANet (ResNet101) is not as good as that of DANet (ResNet50), this may be because the number of DANet (ResNet101) parameters is too large for this data set, which will lead to a slight overfitting phenomenon and thus affect the detection results of the model.

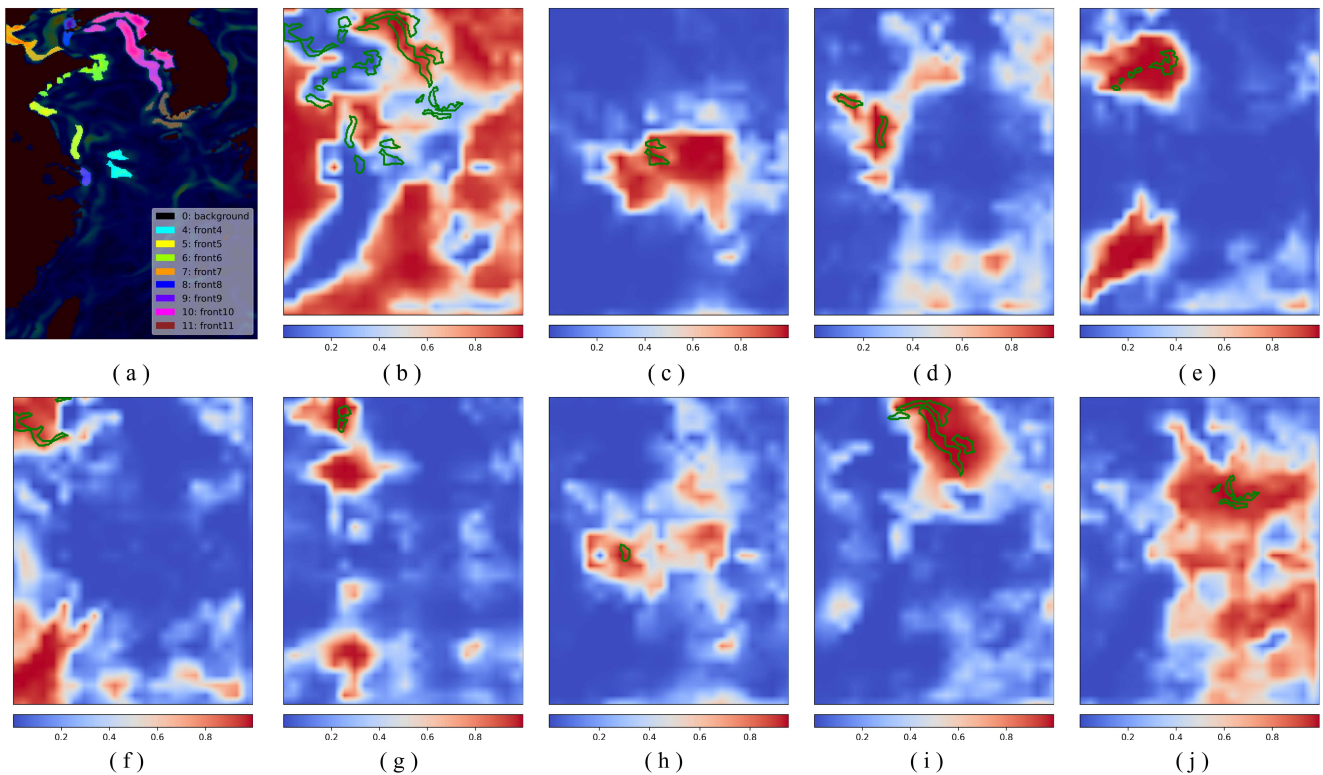

Fig. 6. Attention maps of one day in June, the green curve in the figure shows the edge of the ocean front. (a) ground truth. (b) background. (c) front4. (d) front5. (e) front6. (f) front7. (g) front8. (h) front9. (i) front10. (j) front11.

TABLE III
RESULTS OF ABLATION EXPERIMENT ON CSU AND LA

| Model | Seasonal Encoding | Location Encoding | mIoU |
|---|---|---|---|
| Basenet | × | × | 68.98 |
| Basenet + CSU | × | × | 70.11 |
| Basenet + CSU | Season | × | 70.59 |
| Basenet + CSU | Month | × | 70.60 |
| Basenet + LA | × | × | 69.54 |
| Basenet + LA | × | CoordConv | 70.15 |
| Basenet + LA | × | Positional Encoding 2D | 70.26 |

TABLE IV
RESULTS OF ABLATION EXPERIMENTS ON DATA AUGMENT

| Method | Data Augment | mIoU |
|---|---|---|
| Ours |  | 68.86 |
| Ours | √ | 71.25 |

In addition, we visualized the detection results (Fig. 5). Compared with other models, we can clearly see the superiority of the detection results of our model, especially for front3 (at the bottom of the image), front4 (in the middle of the image) and front7 (at the top left corner of the image).

C. Ablation Experiment

Ablation experiments were carried out to demonstrate the validity of channel supervision unit (CSU) and location attention mechanism (LA). In this experiment, we define the Basenet to represent the network removal of CSU and LA in our model. For CSU test, two seasonal feature encoding methods are designed, one-hot encoding by month (feature vector of length 12) and one-hot encoding by season (feature vector of length 4). For LA test, except for using the positional encoding 2D, the CoordConv [31] are also used for location encoding. Compared with positional encoding 2D, the implementation of CoordConv is relatively simple and fast, which only needs to first concatenate the horizontal and vertical coordinates of the pixel channel-wise to the feature map, and then carry out a convolution operation. Table III shows the results of ablation experiment. On the whole, both CSU and LA have improved the detection accuracy of model, and the improvement of CSU is greater than that of LA. Moreover, the adding seasonal and location encoding is better than the method without any encoding. For CSU, there is little difference between encoding methods by season and by month. For LA, positional encoding 2D produces slightly better results than the CoordConv's.

TABLE V
COMPARISON WITH WEIN

| Model | background | front | mIoU |
|---|---|---|---|
| WEIN | **98.61** | 38.88 | 68.75 |
| Ours | 98.58 | **77.16** | **87.87** |

TABLE VI
BINARY DETECTION RESULTS USING WEIN'S TEST SET SELECTION STRATEGY

| Model | background | front | mIoU |
|---|---|---|---|
| WEIN | 98.80 | 44.93 | 71.86 |
| Ours | **98.81** | **80.95** | **89.88** |

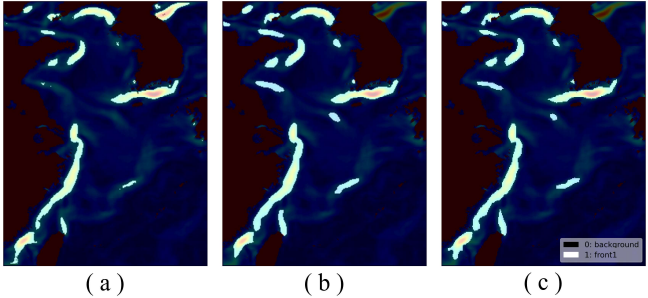

Fig. 7. Comparison with WEIN by binary classification. (a) WEIN. (b) our proposed model. (c) ground truth.

In addition, we also conducted an ablation experiment for data augment (see Table IV), and the experimental results confirmed the effectiveness of data augment.

### D. Visualization of Location Attention Mechanism

Ocean fronts have obvious seasonality. In spring and winter, there are more classes of ocean fronts and they are widely distributed. While in summer and autumn, there are fewer classes of ocean fronts and they are mainly distributed in the Bohai Sea, Yellow Sea and near Korea as is shown in the upper part of Fig. 6(a). To further demonstrate the effectiveness of location attention mechanism, we visualize the upsampled attention maps of fronts for each class in summer or autumn (Fig. 6), spring or winter (in the appendix) respectively, and plot the edges of each front on their own attention maps for the convenience of observation. In Fig. 6, each pixel in the map is a weight that ranges from 0 to 1, which can be considered as the probability value of the current class of fronts occurring at this pixel. The red area in Fig. 6 is the high-weight area, which indicates the area more likely to appear ocean front of the target class or background, while the blue area means the low-weight area and is less likely to appear ocean front of the target class or background. It can be seen that except for the misjudgment of front10 in the background attention map (Fig. 6(b)), other fronts are given higher weights in the sea area where fronts often occur. These visualization results further show that our model does capture the relationship between the ocean front and its sea area location.

### E. Comparison with the Existing Ocean Front Detection Model

It is worth mentioning that the existing end-to-end ocean front detection model does not carry out multi-class detection, and only the model WEIN [5] can achieve binary-class ocean front detection at pixel-level. In order to compare our model with WEIN for the binary classification of ocean front, we convert the 12 classes into 2 classes by setting the class label of all pixels contained in the ocean fronts to 1. The comparison results in Table V shows that our multi-class detection model still achieves state-of-the-art performance in binary classification of ocean fronts. Compared with WEIN, the detection accuracy of ocean front has been greatly improved by 38.28%. Fig. 7 visualizes the detection results compared with WEIN, it can be intuitively seen that WEIN has a serious problem of missing detection of ocean fronts.

In addition, we adopt WEIN's test set selection strategy to randomly select 5 days from each month as test set (a total of 240 samples) and train two models respectively. From the results showed in Table VI, we find that the detection accuracy of both models has been improved by about 2%-3%. Moreover, we also use WEIN's test set selection strategy to perform multi-class detection on our model, and got the satisfied detection accuracy with the mIoU of 77.31 (about 6% improvement).

We also adopt another random test set selection strategy. We randomly select 90 days from each year as the test set (a total of 360 samples). And the model achieved similar and good results as WEIN's test set selection strategy above. We found that once the date of the test sample is close to the date of the sample in the train set, the detection accuracy will be greatly improved. This phenomenon indicates the temporal continuity of the ocean front evolution. In fact, the occurrence, development and extinction of ocean fronts usually last for many days. As is shown in Fig. 8, the ocean fronts of a random week in 2018 are visualized, clearly showing the similarity of shapes during the evolution of ocean fronts (for example, the westward convergence of front3 and southward extension of front5). Therefore, in the future, temporal continuity can be introduced through a specific network structure to further improve the detection accuracy of our model.

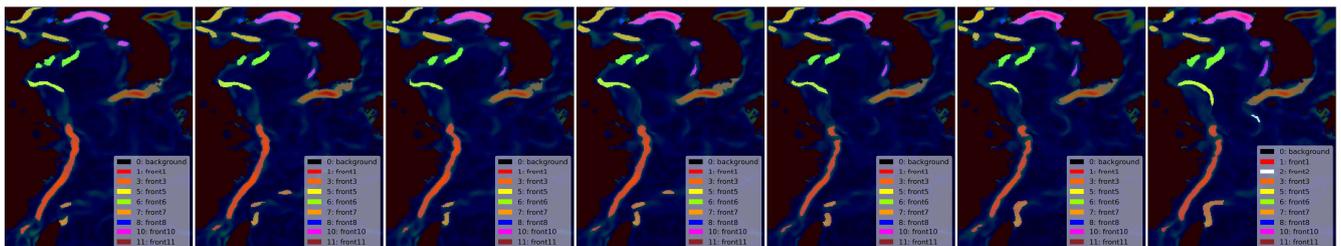

Fig. 8. Visualization of weekly distribution of ocean fronts. The dates from left to right are December 20, 2018 to December 26, 2018.

## VI. Conclusion

In this paper, we propose a semantic segmentation model called LSENet for multi-class detection of ocean fronts at pixel-level. In this model, we consider the different physical characteristics of ocean fronts, and introduce the seasonality and spatial dependence through the channel supervision unit structure and the location attention mechanism respectively for improving the detection accuracy. We carry out many comparison experiments on our own data sets and demonstrate the superiority of our model and the unique designed structure. In the future, we will integrate the temporal continuity of ocean front into our model by a new network structure.

## Appendix

Fig. 9 gives the implementation details of location attention mechanism.

Fig. 10 shows the attention maps in spring or winter, which includes all 11 fronts. In the attention map of each front, the sea area where front frequently occurs is basically given a higher weight. Additionally, the front10 in the background attention map (Fig. 10(b)) is misjudged as in Fig. 6. However, the mechanism behind this phenomenon is still unclear and needs in-depth study.

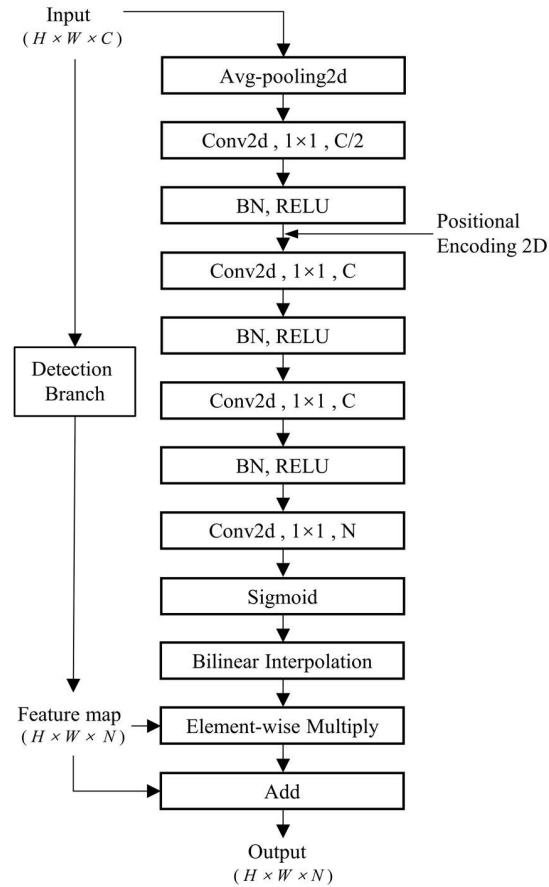

Fig. 9.  Implementation details of location attention mechanism.

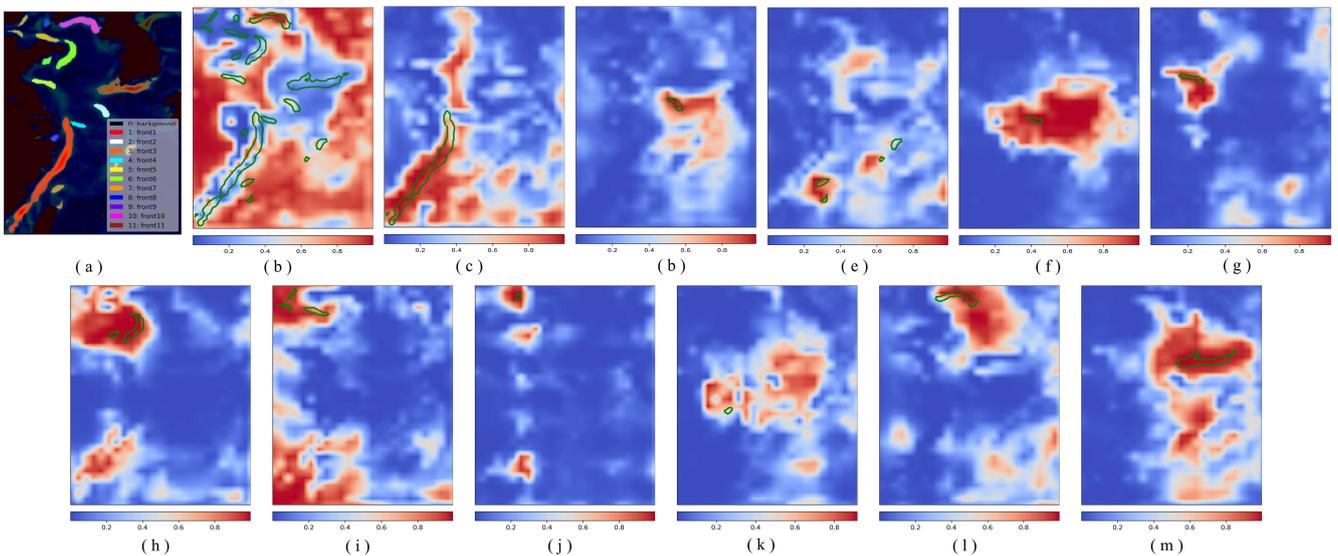

Fig. 10.  Attention maps of one day in January, the green curve in the figure shows the edge of the ocean front. (a) ground truth. (b) background. (c) front1. (d) front2. (e) front3. (f) front4. (g) front5. (h) front6. (i) front7. (j) front8. (k) front9. (l) front10. (m) front11.